# Exploration of RNA Editing and Design of Robust Genetic Algorithms

**Chien-Feng Huang**
Modeling, Algorithms and Informatics Group (CCS-3)
Computer and Computational Sciences Division
Los Alamos National Laboratory, MS B256
Los Alamos, NM 87545, USA
cfhuang@lanl.gov

**Luis M. Rocha**
Modeling, Algorithms and Informatics Group (CCS-3)
Computer and Computational Sciences Division
Los Alamos National Laboratory, MS B256
Los Alamos, NM 87545, USA
rocha@lanl.gov

**Abstract -- This paper presents our computational methodology using Genetic Algorithms (GA) for exploring the nature of RNA editing. These models are constructed using several genetic editing characteristics that are gleaned from the RNA editing system as observed in several organisms. We have expanded the traditional Genetic Algorithm with artificial editing mechanisms as proposed by (Rocha, 1997). The incorporation of editing mechanisms provides a means for artificial agents with genetic descriptions to gain greater phenotypic plasticity, which may be environmentally regulated. Our first implementations of these ideas have shed some light into the evolutionary implications of RNA editing. Based on these understandings, we demonstrate how to select proper RNA editors for designing more robust GAs, and the results will show promising applications to real-world problems. We expect that the framework proposed will both facilitate determining the evolutionary role of RNA editing in biology, and advance the current state of research in Genetic Algorithms.**

## 1 Introduction

The most famous RNA editing system is that of the African Trypanosomes (Benne, 1993; Stuart, 1993). Its genetic material was found to possess strange sequence features such as genes without translational initiation and termination codons, frame shifted genes, etc. Furthermore, observation of mRNA's showed that many of them were significantly different from the genetic material from which they had been transcribed. These facts suggested that mRNA's were edited post-transcriptionally. It was later recognized that this editing was performed by guide RNA's (gRNA's) coded mostly by what was previously thought of as non-functional genetic material (Sturn and Simpson, 1990). In this particular genetic system, gRNA's operate by inserting, and sometimes deleting, uridines. To appreciate the effect of this edition let us consider Figure 1. The first example (Benne, 1993, p. 14) shows a massive uridine insertion (lowercase u's); the amino acid sequence that would be obtained prior to any edition is shown on top of the base sequence, and the amino acid sequence obtained after edition is shown in the gray box. The second example shows how, potentially, the insertion of a single uridine can change dramatically the amino acid sequence obtained; in this case, a termination codon is introduced. It is important to retain that a mRNA molecule can be more or less edited according to the concentrations of the editing operators it encounters. Thus, several different proteins coded by the same gene may coexist in an organism or even a cell, if all (or some) of the mRNA's obtained from the same gene, but edited differently, are meaningful to the translation mechanism.

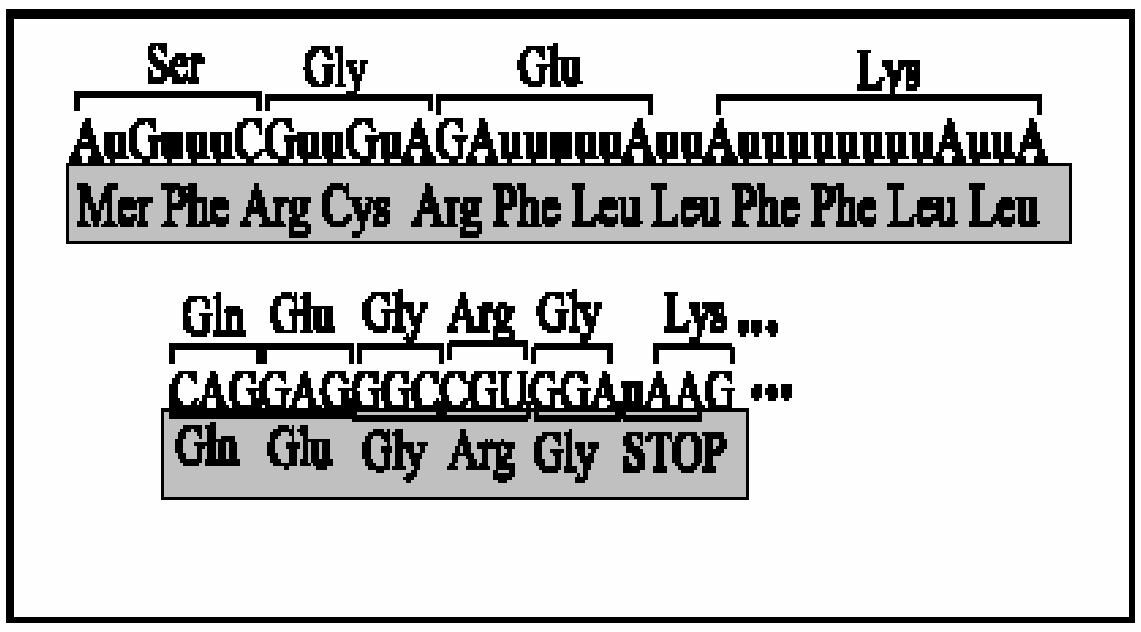

**Figure 1. U-insertion in Trypanosomes' RNA**

The role of RNA editing in the development of more complex organisms has also been shown to be important. Lomeli et al. (1994) discovered that the extent of RNA editing affecting a type of receptor channels responsible for the mediation of excitatory postsynaptic currents in the central nervous system, increases in rat brain development. As a consequence, the kinetic aspects of these channels differ according to the time of their creation in the brain's developmental process. Another example is that the development of rats without a gene (ADAR1) known to be involved in RNA editing, terminates midterm (Wang et al., 2000). This showed that RNA Editing is more prevalent and important than previously thought. RNA editing processes have also been identified in mammalian brains (Simpson and Emerson, 1996), including human brains (Mittaz et al., 1997).

Although RNA editing seems to play an essential role in the development of some genetic systems and more and more editing mechanisms have been identified, not much has been advanced to understand the potential evolutionary advantages, if any, that RNA editing processes may have provided. To acquire insights for answering this question, we need a systematic study on how RNA editing works. Furthermore, a deeper understanding of the nature of RNA editing can be exploited to improve evolutionary computation tools and their applications to complex, real-world problems. This paper reports some of our results towards these two goals.

## 2 Introducing Editing in Genetic Algorithms

In science and technology Genetic Algorithms (GA) (Holland, 1975) have been used as computational models of natural evolutionary systems and as adaptive algorithms for solving optimization problems. Although GA are simplified, idealized models of evolutionary systems, this approach has led to important discoveries in both natural and artificial systems. For instance, Lindgren (1991) used GA to evolve iterated prisoner's dilemma rules and modeled many processes observed in biological evolution such as stasis, punctuated equilibria, varying speeds of evolution, mass extinctions, symbiosis, and complexity increase. Since GA, at the very least, are a good model of adaptive processes obtained by variation and selection of genotypes in natural systems, in the present work we use them to explore the evolutionary implications of edited genotypes, such as the RNA editing system. Table 1 depicts the process of a simple genetic algorithm.

Table 1. Mechanism of a simple GA

1. Randomly generate an initial population of $l$ $n$-bit agents, each defined by a genotype string (chromosome) of symbols from a small alphabet.
2. Evaluate each agent's (phenotype) fitness.
3. Repeat until $l$ offspring agents have been created.
   a. select a pair of parent agents for mating;
   b. apply crossover operator to genotype string;
   c. apply mutation operator to genotype string.
4. Replace the current population with the new population.
5. Go to Step 2 until terminating condition.

GAs operate on an evolving population of artificial organisms, or agents. Each agent is comprised of a genotype and a phenotype. Evolution occurs by iterated stochastic variation of genotypes, and selection of the best phenotypes in an environment according to a fitness function. In machine learning, the phenotype is a candidate solution to some optimization problem, while the genotype is an encoding, or description, of that solution by means of a domain independent representation, namely, binary symbol strings (or chromosomes). In traditional GAs, this code between genotype and phenotype is a direct and unique mapping. In biological genetic systems, however, there exists a multitude of processes, taking place between the transcription of a description and its expression, responsible for the establishment of an uncertain relation between genotype and phenotype. For instance, it was shown that RNA editing has the power to dramatically alter gene expression (Pollack, 1994, P. 78): "cells with different mixes of (editing mechanisms) may edit a transcript from the same gene differently, thereby making different proteins from the same opened gene."

In other words, in a genetic system with RNA editing, before a gene is translated into the space of proteins it may be altered through interactions with other types of molecules, namely RNA editors such as gRNA's. Based upon this analogy, (Rocha, 1995; Rocha, 1997) proposed a new class of GAs that implement a process of stochastic edition of the genotypes (chromosomes) of agents, prior to being translated into phenotypes (candidate solutions). The editing process is implemented by a set of editors with different editing functions, such as insertion or deletion of symbols in the original chromosomes. Before chromosomes can be translated into the space of solutions, they must "pass" through successive layers of editors, present in different concentrations. In each generation, each chromosome has a certain probability (given by the concentrations) of encountering an editor in its layer. If an editor matches some subsequence of the chromosome when they encounter each other, the editor's function is applied and the chromosome is edited. The detailed implementation of the simplest GA with Edition (GAE) is described in the following:

The GAE model consists of a family of $r$ $m$-bit strings, denoted as ($E_1, E_2$, …, $E_r$), which is used as the set of editors for the chromosomes of the agents in a GA population. The length of the editor strings is assumed much smaller than that of the chromosomes: $m << n$, usually an order of magnitude. An editor $E_j$ is said to match a substring, of size $m$, of a chromosome, $S$, at position $k$ if $e_i = s_{k+i}$, $i$=1,2, …, $m$, 1 = $k$ = $n$-$m$, where $e_i$ and $s_i$ denote the $i$-th bit value of $E_j$ and $S$, respectively. For each editor, there exists an associated editing function that specifies how a particular editor edits the chromosomes: when the editor matches a portion of a chromosome, a number of bits are inserted into or deleted from the chromosome.

For instance, if the editing function of editor $E_j$ is to add one randomly generated allele at $s_{k+m+1}$ when $E_j$ matches $S$ at position $k$, then all alleles of $S$ from position $k+m+1$ to $n-1$ are shifted one position to the right (the allele at position $n$ is removed). Analogously, if the editing function of editor $E_j$ is to delete an allele, this editor will instead delete the allele at $s_{k+m+1}$ when $E_j$ matches $S$ at position $k$. All the alleles after position $k+m+1$ are shifted in the inverse direction (one randomly generated allele is then assigned at position $n$).

Finally, let the concentration of the editor family be defined by ($v_1, v_2$, …, $v_r$). This means that the concentration of editor $E_j$ is denoted as $v_j$, and the probability that $S$ encounters $E_j$ is thus given by $v_j$. With these settings, the algorithm for the GA with string editing is essentially the same as the regular GA, except that step 2 in Table 1 is now more complicated and redefined as:

"For each individual in the GA population, apply each editor $E_j$ with probability $v_j$ (i.e., concentration). If $E_j$ matches the individual's chromosome *S*, then edit *S* with the editing function associated with $E_j$ and evaluate the resulting individual's fitness."

# 3 Properties of Genotype Editing

## 3.1 Improvement Rate and Building-Block Dynamics

How rapid is evolutionary change, and what determines the rates, patterns, and causes of change, or lack thereof? Answers to these questions can tell us much about the evolutionary process. The study of evolutionary rate in the context of GA usually involves defining a performance measure that embodies the idea of rate of improvement, so that its change over time can be monitored for investigation.

In many practical problems, a traditional performance metric is the "best-so-far" curve that plots the fitness of the best individual that has been seen thus far by generation *n*. To understand how Genotype Editing works in the GAE model, we employ a testbed, the small Royal Road S1, which is a miniature of the class of the "Royal Road" functions (Forrest and Mitchell, 1993).

**Table 2. Small royal road function S1**

$s_1$ = 11111***********************************; $c_1 = 10$
$s_2$ = *****11111******************************; $c_2 = 10$
$s_3$ = **********11111*************************; $c_3 = 10$
$s_4$ = ***************11111********************; $c_4 = 10$
$s_5$ = ********************11111***************; $c_5 = 10$
$s_6$ = *************************11111**********; $c_6 = 10$
$s_7$ = ******************************11111*****; $c_7 = 10$
$s_8$ = ***********************************11111; $c_8 = 10$

Table 2 illustrates the schematic of the small Royal Road function S1. This function involves a set of schemata $S = (s_1, ..., s_8)$ and the fitness of a bit string (chromosome) *x* is defined as

$$F(x) = \sum_{s_i \in S} c_i s_{s_i}(x),$$

where each $c_i$ is a value assigned to the schema $s_i$ as defined in the table; $s_{s_i}(x)$ is defined as 1 if *x* is an instance of $s_i$ and 0 otherwise. In this function, the fitness of the global optimum string (40 1's) is 10*8 = 80.

We select this Royal Road function as a testbed because it belongs to a class of building-block-based functions, in which search advancements depend entirely on the discovery and exploitation of building blocks. This serves as an idealized testbed for observing how editing improves the GA's search power by tracing the origin of each advance in performance.

The GAE experiments conducted in this subsection are based on a binary tournament selection, one-point crossover and mutation rates of 0.7 and 0.005, respectively; and population size 40 over 200 generations for 50 runs. A family of 5 editors is randomly generated, with editor length selected in the range of 2 to 4 bits. Table 3 shows the corresponding parameters generated for these editors.

**Table 3. Parameters of the five RNA editors**

| | editor 1 | editor 2 | editor 3 | editor 4 | editor 5 |
|---|---|---|---|---|---|
| length | 4 | 4 | 4 | 2 | 4 |
| alleles | {1,1,1,0} | {0,0,1,1} | {0,1,0,1} | {0,0} | {0,1,1,1} |
| concentration | 0.0635 | 0.0476 | 0.7302 | 0.2857 | 0.3175 |
| function | delete 4 bits | add 3 bits | delete 1 bit | delete 3 bits | delete 2 bits |

In Table 3, "length", "alleles" and "concentration" denote the length, alleles and concentration of each editor, respectively; and "function" denotes the corresponding editing function. For example, the editing function of editor 1 is to delete 4 bits, meaning that this editor deletes 4 alleles at the positions following the chromosome substring that matches the editor.

The empirical results are displayed in Figure 2. One can see that the averaged best-so-far located by the GA with editors is 80 at the end of the experiments,[1] indicating that the optimum has been found by the GA for all 50 runs. On the other hand, our detailed results show that in the case of the GA without editors the optimum is located in 17 out of 50 runs, and the averaged best-so-far only reaches fitness of around 70 by 200 generations.

A microscopic inspection shows that the search power of the regular GA is limited by the effects of hitchhiking: a well-known phenomenon that occurs when some newly discovered allele (or sets of alleles) offers great fitness advantages. As that allele spreads quickly through the population, the closely linked alleles (though they may make no contribution to the fitness) could propagate to the next generation by hitchhiking on that allele. The rapid occupancy of those non-relevant alleles thus greatly reduces exploration of alternatives at those loci. They either drown out other already-discovered alleles that are advantageous, or leave no room for not-yet-discovered beneficial alleles.

In GA research, hitchhiking has been identified as a major problem that limits implicit parallelism by reducing the sampling frequency of various building blocks (Forrest and Micthell, 1993). We can trace hitchhiking directly by plotting the densities (percentage of the population that are instances) of the relevant schemata over time to observe how editing suppresses hitchhiking.

Figure 3 is a typical run that illustrates such density dynamics. One can see that the optimum has never been found in the GA without editors, because $s_3$ was never discovered. A closer inspection shows that some hitchhikers, 11010, of $s_2$ at the location of $s_3$ indeed preclude the discovery of $s_3$. However, in the GA with editors, the editors tend to edit the hitchhikers by matching subsequences of $s_2$ and offer a larger likelihood for the GA to discover $s_3$. This demonstrates how the editing mechanism can improve the GA's search performance by suppressing the effects of hitchhiking.

[1] The value of the averaged best-so-far performance metric is calculated by averaging the best-so-fars obtained at each generation for all 50 runs, where the vertical bars overlaying the metric curves represent the 95-percent confidence intervals. This applies to all the experimental results obtained in this paper.

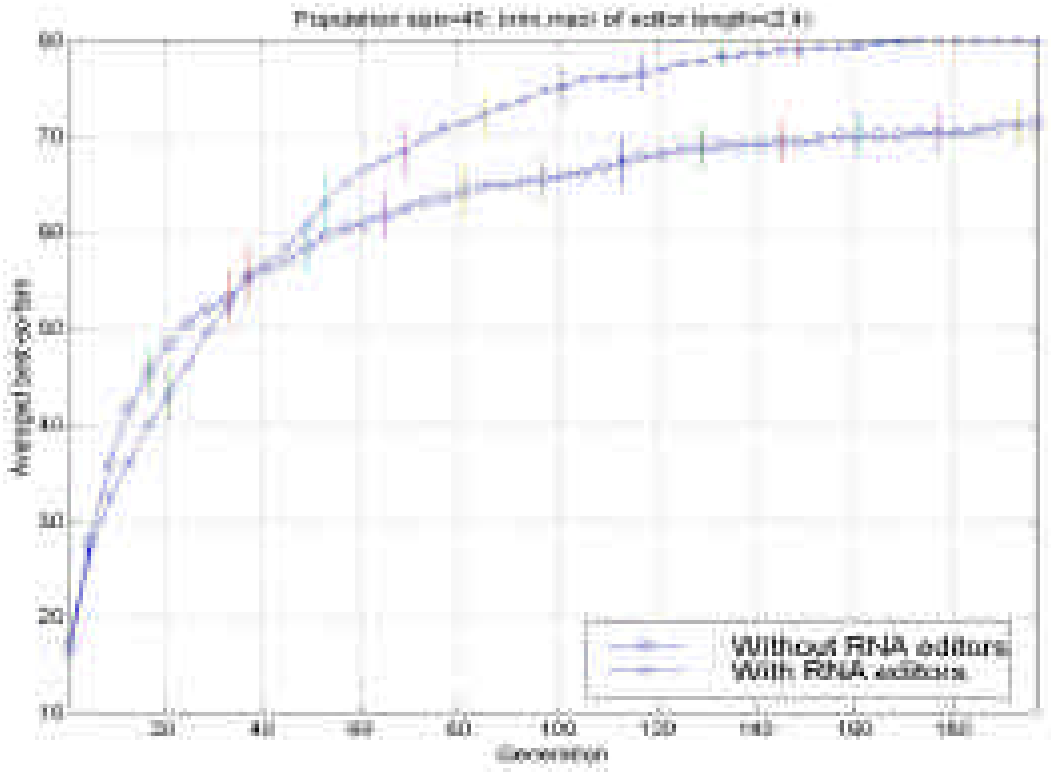

**Figure 2. Averaged best-so-far performance**

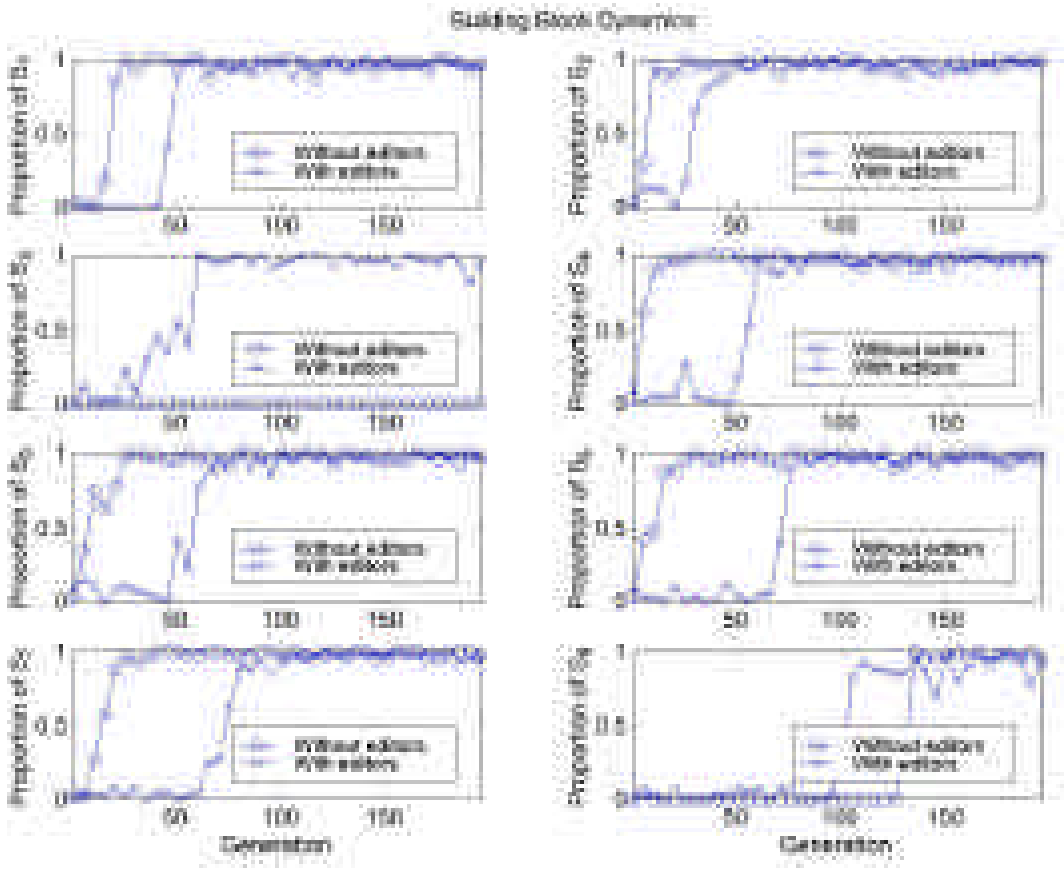

**Figure 3. Building Block Dynamics**

### 3.2 Effects of Size of the Family of Editors

An important parameter that may play a key role in the GA's search power is the size of the family of editors. To study the effects of this factor we conduct experiments using two different families of two and ten editors (the other parameters are generated as in the preceding subsection), in comparison with the family of five editors studied previously. Figure 4.a displays the experimental results for 50 runs, in which the GAE with five editors outperforms the other two GAEs. (The optimum is found in all the 50 runs for the GAE with five editors, but not in the other two GAEs.)

Further results on editing frequency -- the total number of times all editors edited chromosomes in a generation -- illustrated in Figure 4.b show that, in the beginning of the experiments, the editing frequency for the GAE with two or five editors is substantially smaller than that of the GAE with ten editors. These results are quite intuitive, since more editors tend to incur more editing processes. Furthermore, in case of the GAE with five editors, the most striking difference is that the corresponding editing frequency declines dramatically as the GAE's population evolves, and tends to drop to zero at the end of the experiments.

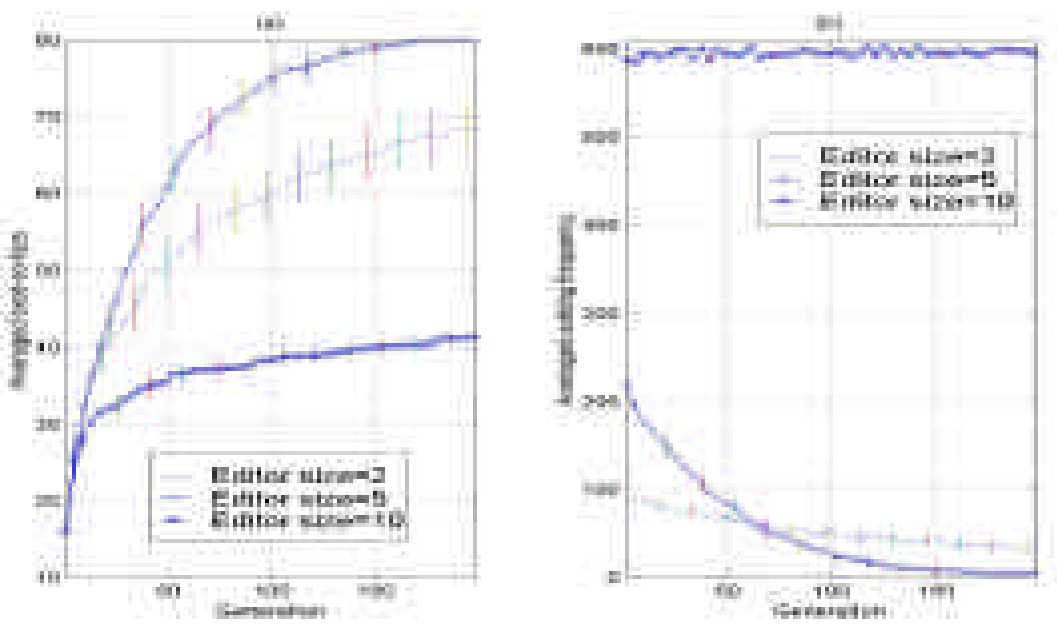

**Figure 4. Effects of size of the editor family**

To further elucidate the effects of size of the editor family, Figure 5.a displays results of editing frequency in a typical run for each type of GAE. (The corresponding maximal fitness located by the GAE with two, five, and ten editors is 70, 80 and 50, respectively.) One can notice that in the typical GAE runs where the optimum is not found (i.e., the cases for 2 and 10 editors), the editing frequency does not significantly drop to zero near the end of the experiments. It appears that these GAs' populations continue utilizing the editors to explore the search space. This is the reason that the corresponding population diversity displayed in Figure 5.b[2] is far from zero in the case of the GAE with 10 editors. For the GAE with 2 editors, the best-so-far fitness located is close to the optimum -- the results in Figure 5.a and 5.b show that the degree of editing is then reduced and the population is not as diverse as that of the GAE with 10 editors. All this indicates that the system settles into a dynamic equilibrium in which the exploratory power of the editing process is balanced by the exploitative pressure of selection.

For the case of the GA with 5 editors, the results displayed in Figure 5, however, show that the GA's population diversity is lost and the editing process ultimately comes to an end. Based on the effects of editor length and concentration, in the next two subsections we will present more results to support our observation.

---

[2] To measure diversity at the *i-th* locus of a GA string, a simple bitwise diversity metric is defined as (Mahfoud, 1995): $D_i = 1 - 2|0.5 - p_i|$, where $p_i$ is the proportion of 1s at locus *i* in the current generation. Averaging the bitwise diversity metric over all loci offers a combined allelic diversity measure for the population: $D = (\sum_{i=1}^{l} D_i)/l$. *D* has a value of 1 when the proportion of 1s at each locus is 0.5 and 0 when all of the loci are fixed to either 0 or 1. Effectively it measures how close the allele frequency is to a random population (1 being closest).

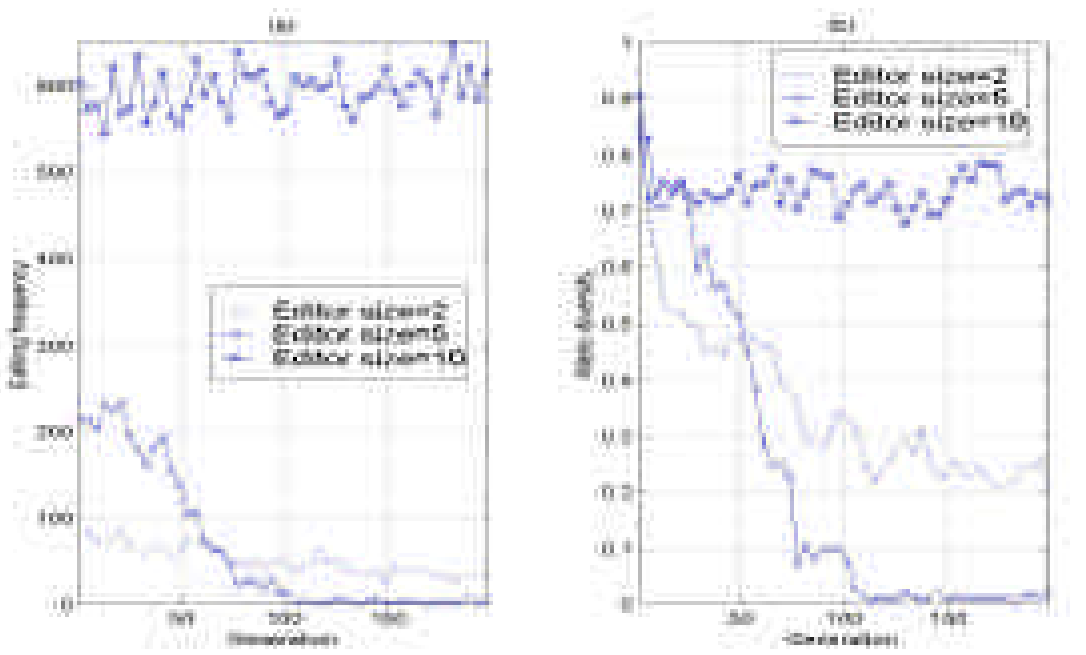

**Figure 5 Editing frequency and population diversity**

### 3.3 Effects of Editor Length

Another important parameter that may also play a critical role is the length of editors. To examine the effects of this factor we conduct experiments for another two GAEs where all the five editors are of 2 bits or 10 bits. (All the other parameters are generated by the same way used in Section 3.1.) Figure 6.a illustrates the results for these GAEs, in which the GAE of Section 3.1 (denoted as “Editor length=2...4” in the figure) outperforms the other two GAEs. Our hypothesis is that, as the length of editors is too long, matchings between editors and subsequences of the GAE’s chromosomes are rather unlikely, thereby inducing almost no editing. On the contrary, if the length of editors is too short, numerous matchings may occur and the GAE’s population will undergo considerable editing processes. This may result in serious disruptive effect on fit individuals.

In other words, the performance discrepancy of the GAEs with different editor length may again depend on editing frequency. The empirical results for editing frequency shown in Figure 6.b confirm our expectation. The editing frequency for the GAE with 10-bit editors (nearly zero frequency over the whole course of the experiments) is far smaller than that of the GAE with 2-bit editors. In this case, the editors make almost no contribution to the GAE’s search power. Nonetheless, for the GAE with 2-bit editors, it is obvious that the GAE undergoes considerable editing processes which in turn disrupt the already-discovered fit individuals.

As for the GAE used in Section 3.1 (with editors of 2 to 4 bits long), the results show that the GA’s population undergoes moderate editing processes in the beginning of the evolutionary process, which seems to facilitate the GAE’s exploration of the search space. Therefore, proper length of the editors is essential to achieve search benefits, and a beneficial editing mechanism would require moderate editing frequency.

### 3.4 Effects of Editor Concentration

As a further illustration for the Royal Road testbed, we examine the effects of editor concentration on the GAE’s search performance. Instead of various concentrations of the editors used in Section 3.1, each editor is now given concentration of 1, meaning that the probability that the chromosomes encounter each editor is 1. Figure 7.a and 7.b display the effects of concentration and the corresponding editing frequency. (Concentration 1 and 2 in the figure denote the concentration used in Section 3.1 and this subsection, respectively.) Since the probability of the chromosomes meeting with editors is now 1, the population would naturally undergo more editions than in the GAE with smaller editor concentrations.

These results again indicate that the performance difference lies in the number of the performed editions. As the GA’s population is considerably edited by the editors, too much exploration of the search space would then generate deleterious effects on performance advancement. Appropriate editor concentration is thus essential for the GAE, since a beneficial edition requires proper editor’s concentration to induce moderate editing processes.

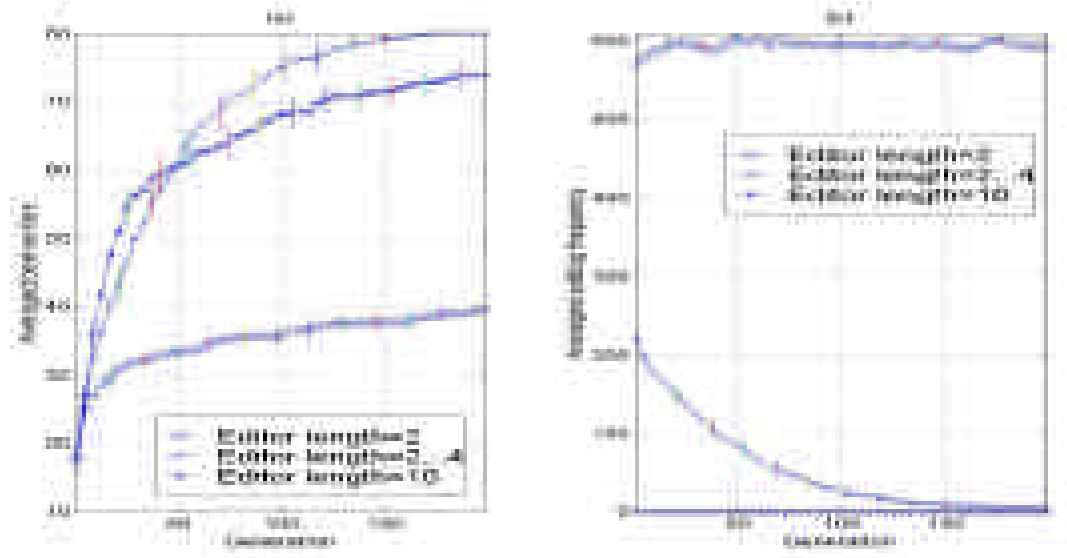

**Figure 6. Effects of editor length**

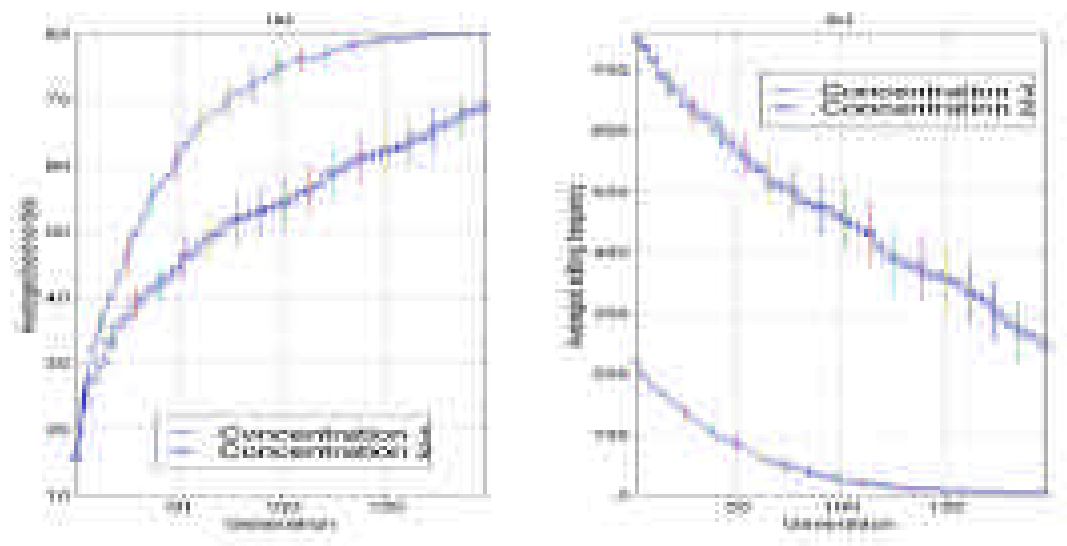

**Figure 7. Effects of editor concentration**

### 3.5 Effects of Editor Function

As the last illustration, we examine the effects of editor function on the GAE’s search performance. Instead of the editor functions used in Section 3.1, the functions of all the editors are now designated to delete 10 bits, meaning that the chromosomes will encounter massive gene deletions when they are matched by the editors. Figure 8.a and 8.b display the effects of the editor functions and the corresponding editing frequency. (Function 1 and 2 in the figure denote the editor functions used in Section 3.1 and this subsection, respectively.) Since the gene deletion frequency of the chromosomes is now increased, the GAE’s population would naturally undergo more disruptive processes than the GAE used in Section 3.1.[3]

These results indicate that the performance difference lies in the degree of gene deletion in chromosomes. As the editors remove considerabe genes of chromosome, beneficial genes tend to be deleted, which would in turn hamper the GAE's search. Appropriate editor function is thus crucial for the GAE to gain subtstantial search progress.

---

[3] We have obtained similar results for massive gene insertions (not shown in this paper).

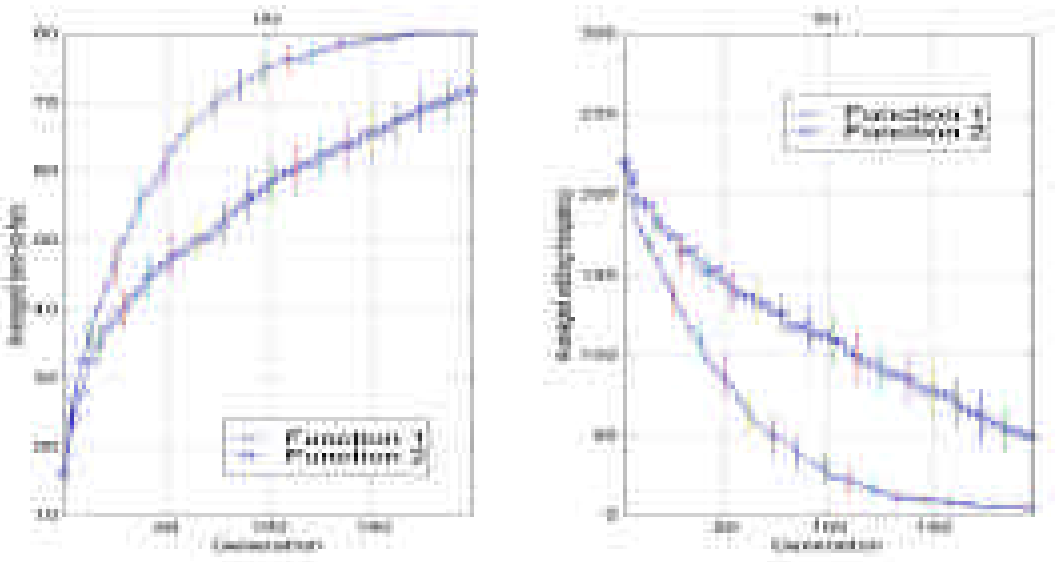

**Figure 8. Effects of editor function**

## 4 Applications

The study of Genotype Editing has provided us with insights into how to choose editor parameters for developing more robust GAs. Basically, in order to faciliate the GAE's search process, the guidelines are:

- the size of the editor family, the length and concentration of the editors need to be moderate so as to avoid over or under-editing processes;
- the editor function should be far from generating massive deletions (or insertions).

Furthermore, the choice of the editor parameters is not absolute, it depends on the problem at hand. In this section we apply these rules to select proper genotype editors for the design of more robust GAEs, and test them on two real, non-building-block-based test functions: an optimal control problem and Michalewicz's epistatic function (Huang, 2002).

### 4.1 An Optimal Control Test Problem

Optimal Control problems often arise in many different fields of engineering and sciences. This class of problems has been well studied from both theoretical and computational perspectives. The models used to describe optimal control problems almost always involve more or less nonlinearity in nature. This often results in the existence of multiple local optima in the area of interest.

In this subsection we employ an artificial optimal control problem designed in (Huang, 2002). The constraints of the artificial optimal control problem are:

$$\frac{d^2z(t)}{dt^2} + sin(z(t))\frac{dz(t)}{dt} + sin(t)cos(z(t))z(t)^3 = sin(t)u_1^2 + cos(t)u_2^2 + sin(t)u_1u_2,$$
$$z(t_0) = 2, \dot{z}(t_0) = 2, t \in [0, 1].$$

The goal is to maximize $z(t_f)^2$ by searching for two constant control variables, $u_1$ and $u_2$ (-5 = $u_1$, $u_2$ = 5). A sketch of this function is displayed on the left side of Figure 8. The X and Y axes represent the index of sample points in parameters $u_1$ and $u_2$ that are used to compute $z(t_f)^2$, which is then plotted on Z-axis. There are clusters of spikes at two corners of the search space, and a hill that occupies most of the space. The magnified view on the right side of Figure 8 shows a clearer view of the height and area of the hill.

As can be seen, the height of the hill is much lower than that of the spikes, but since it occupies most of the search space, we expect that most of the population individuals would be attracted to the hilltop, which then impairs the GA's search power. This problem has been recognized in GA research as premature convergence.

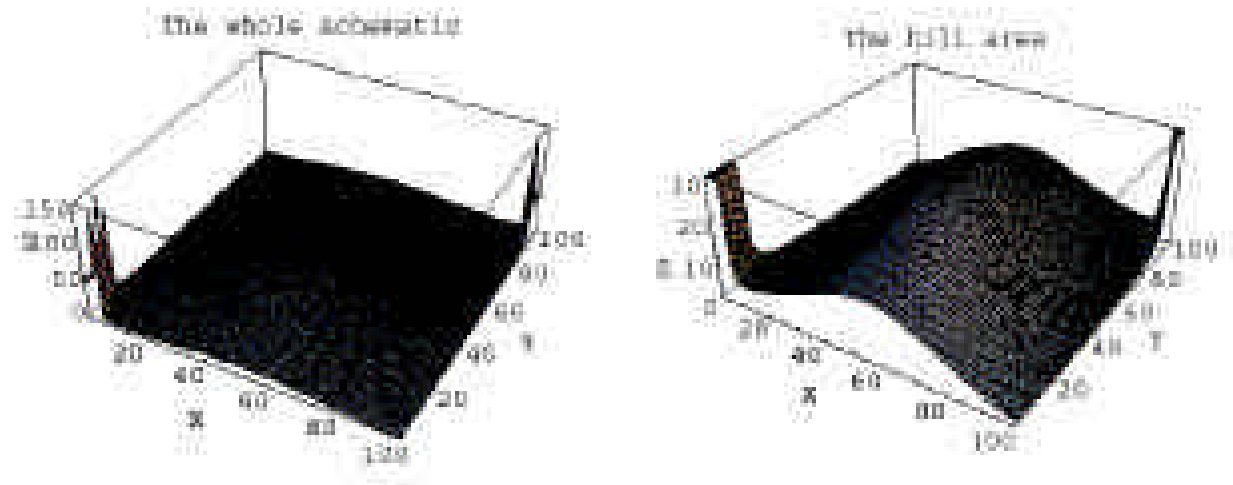

**Figure 9. The optimal control problem**

With the results obtained previously, one may expect that the GAE can use editors to facilitate advance in search by editing the population individuals that prematurely converge on the hill, in order to relocate these individuals into higher fitness spikes. Our main objective in this subsection is to test if the GAE can provide advantages in search.

In this subsection, each of the two variables is encoded by 30 bits, and thus each individual is a binary string of length 60. We use a population size 50, binary tournament selection, and crossover and mutation rates of 0.7 and 0.005, respectively. The experiments are conducted for 100 runs, each run with 200 generations. For the family of editors, since the string length and population size used is of the same order as those used in the last section, we also use five editors, of length between 3 and 6, moderate concentraions, and moderate degree of insertion or deletion. These parameters are shown in Table 4.

**Table 4. Parameters of the five editors**

| | editor 1 | editor 2 | editor 3 | editor 4 | editor 5 |
|---|---|---|---|---|---|
| length | 5 | 4 | 5 | 3 | 6 |
| alleles | {0,0,1,1,0} | {1,0,0,1} | {0,1,1,0,1} | {0,1,1} | {1,1,1,1,0,0} |
| concentration | 0.1410 | 0.7936 | 0.2524 | 0.5885 | 0.0871 |
| function | delete 2 bits | delete 1 bit | add 3 bits | add 2 bits | add 5 bits |

Figure 10 displays the averaged best-so-far performance, which shows that the genotype edition again achieved an advantage in search. As we examine the detailed results, we see that for the case where editors are absent the best-so-far located by the GA is only of 27.01 (the fitness at the hilltop) in nearly 60 out of 100 runs. However, the GAE explores more of the search space and extends the best-so-fars to higher range. This tells us how these editors improve the GAE's search process.

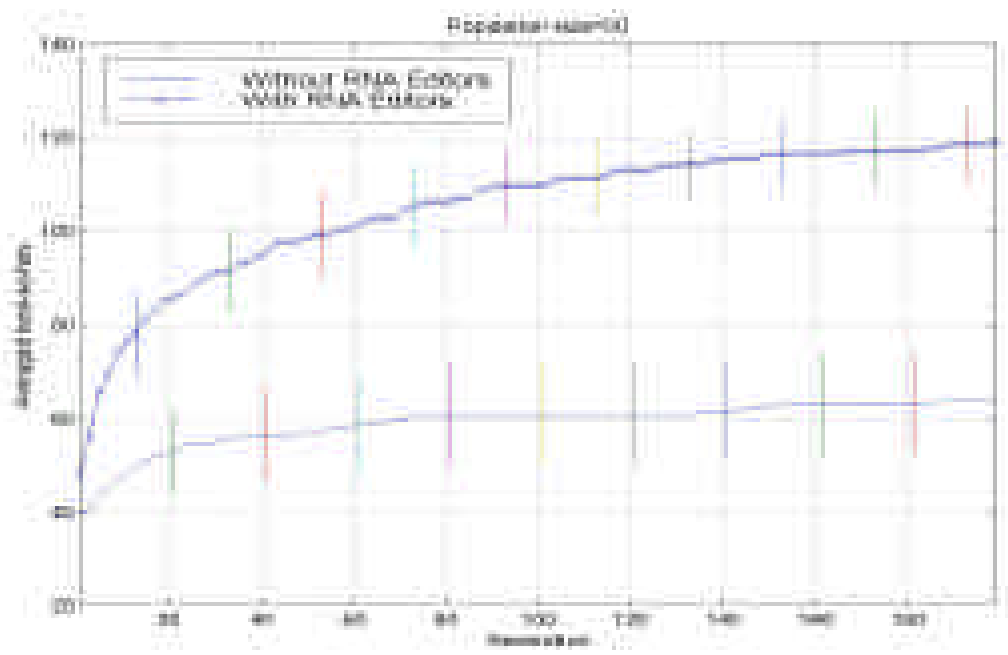

**Figure 10. Averaged best-so-far performance**

### 4.2 Epistatic Michalewicz Function

In contrast to the relatively simple fitness landscape of the optimal control problem, a much more complicated testbed, the modified epistatic Michalewicz function (Huang, 2002), is used:

$$f(\overline{x}) = \sum_{i=1}^{N} sin(y_i) sin^{2m}(\frac{iy_i^2}{\pi}),$$

where

$$y_i = x_i cos\frac{\pi}{6} - x_{i+1} sin\frac{\pi}{6}, \quad \text{if } i \bmod 2 = 1;$$
$$y_i = x_{i-1} sin\frac{\pi}{6} + x_i cos\frac{\pi}{6}, \quad \text{if } i \bmod 2 = 0 \text{ and } i \neq N;$$
$$y_N = x_N,$$

and $m$ = 10, 0 = $x_i$ = π for 1 = $i$ = $N$. A system is little (very) epistatic if the optimal allele for any locus depends on a small (large) number of alleles at other loci. The concept of epistasis in nature corresponds to nonlinearity in the context of GA (Goldberg, 1989).

A sketch of a two-dimensional version of this function is displayed in Figure 11. This function is a highly multimodal, nonlinear and nonseparable testbed. Due to the complicated, nonlinear dependence among alleles, one can expect that this problem presents considerable difficulty to the GA's search.

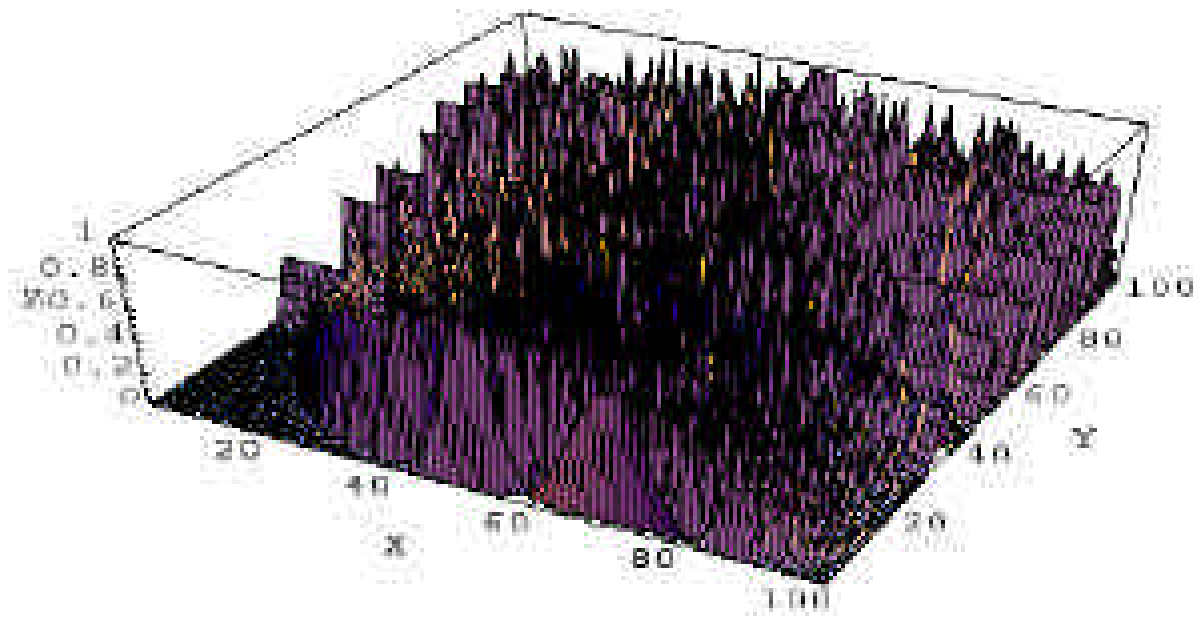

**Figure 11. Modified Michalewicz function**

In this subsection, we use five variables ($N$ = 5), each variable being encoded by 10 bits. Thus each chromosome is a binary string of length 50. The parameters of the 5 editors are:

| | editor 1 | editor 2 | editor 3 | editor 4 | editor 5 |
|---|---|---|---|---|---|
| length | 5 | 5 | 5 | 5 | 5 |
| alleles | {1,1,1,0,0} | {0,1,0,1,1} | {1,1,1,0,1} | {0,1,0,0,0} | {0,0,0,0,0} |
| concentration | 0.762 | 0.54 | 0.254 | 0.159 | 0.159 |
| function | add 1 bit | add 1 bit | add 5 bits | add 3 bits | delete 2 bits |

All other GAE parameter values remain the same as those used in the previous subsection. Figure 12 displays the corresponding averaged best-so-far performance, where one can see that the search performance of the GAE, with the assistance of editors, is improved.

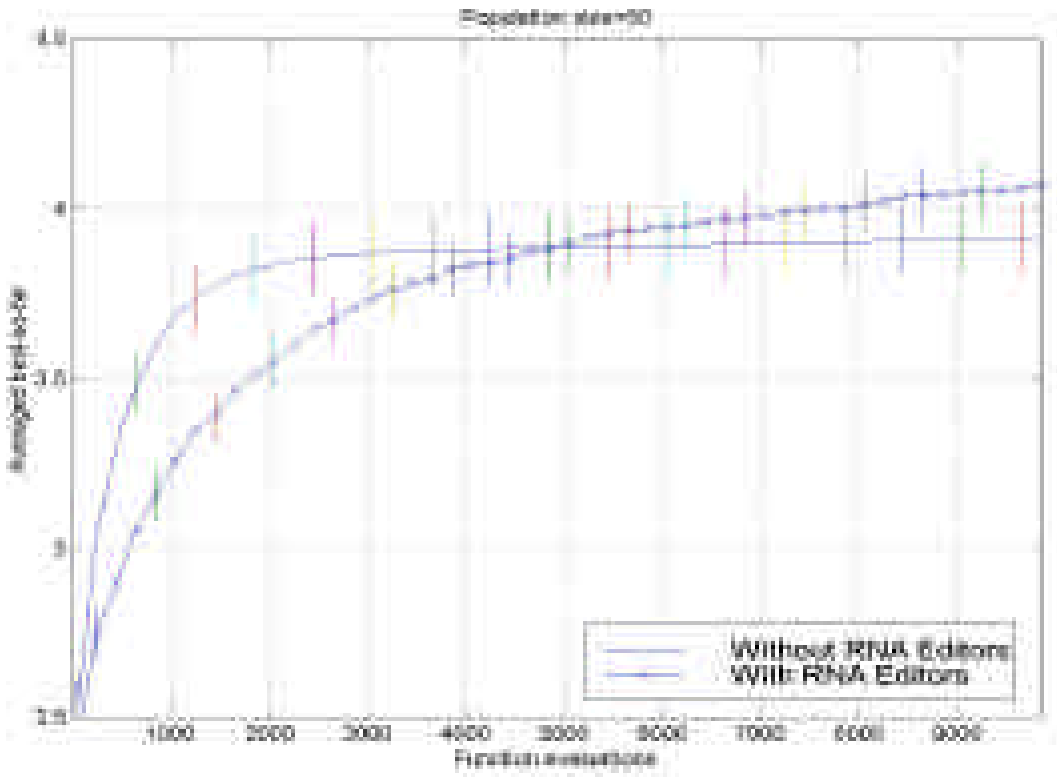

**Figure 12. Averaged best-so-far performance**

## 5 Conclusion and Future Work

We have presented the framework of editing using Genetic Algorithms and tested several evolutionary scenarios. The preliminary results obtained have shed some light into Genotype Editing:

Editing frequency plays a critical role in the evolutionary advantage provided by the editors -- only a moderate degree of editing processes would facilitate organisms' exploration of the search space. Our results also indicate that editing frequency declines dramatically as the population diversity is lost, indicating that the editing process ultimately comes to an end. If the editing frequency does not substantially decrease, the system settles into a dynamic equilibrium where the exploratory power of the editing process is balanced by the exploitative pressure of selection.

We have also learned some rules for setting up editors' parameters to develop robust GAEs. The results obtained on real testbeds show promising applications to practical problems -- in the context of GA the editing mechanisms demonstrate the capability of substantially improving the solution quality in function optimizations and engineering. Together with the insights acquired previously, in future work we aim at conducting more biologically realistic experiments which may lead us towards a better understanding of the advantages of RNA editing in nature, and elaborating the conditions under which the editing framework will result in further improvement in the GA's search performance, as well.

In this paper we discussed GAs with edition solely with constant parameters, such as fixed concentrations, of editors and a stable environment defined by a fixed fitness function. Our preliminary tests (not discussed here), however, also show that constant concentrations of editors may not grant the system any evolutionary advantage when the environment changes. In order to simulate a genetic system in which the linking of editors' concentrations with environmental states may be advantageous in time-varying environments, (Rocha, 1995; Rocha, 1997) proposed a new type of GA known as Contextual Genetic Algorithms (CGA). In this class of algorithms, the concentrations of editors change with the states of the environment, thus introducing a control mechanism leading to phenotypic plasticity and greater evolvability.

We have already constructed a preliminary model that allows the relation between environmental states and editors characteristics (such as concentrations or strings) to be adaptive. Basically, we evolve the concentrations of editors using an additional GA, or allow (slower) mutation of editing strings. This way, editors co-evolve with the population of edited agents in a dynamic environment. Our preliminary results on applying this co-evolving CGA to a simple Royal Road testbed (Huang, 2002) indeed show that as the concentrations of editors co-evolve with edited agents to the environmental demands, the CGA's search performance can be improved with respect to function optimization. We expect that this co-evolved linking of the parameters of editors with changes in environments to be even more powerful in solving dynamic, stochastic real-world problems. Our future work will report on our continued efforts to systematically study and determine the conditions under which CGA can provide artificial agents an improved adaptability to dynamic environments. Such a deeper understanding of CGA will lead us to tackle our two ultimate goals: (1) develop novel evolutionary computation tools for dealing with dynamic real-world tasks, and (2) gain a greater understanding of the evolutionary value of RNA Editing in Biology.